\documentclass[10pt,journal,compsoc]{IEEEtran}
\usepackage{amsmath,amsfonts}
\usepackage{algorithmic}
\usepackage{array}
\usepackage{textcomp}
\usepackage{stfloats}
\usepackage{url}
\usepackage{verbatim}
\usepackage{graphicx}
\usepackage[table]{xcolor}
\usepackage{epsfig}
\usepackage{graphicx}
\usepackage{amsmath}
\usepackage{amssymb}
\usepackage{enumitem}
\usepackage{pifont}

\usepackage{multirow}
\usepackage{tabularx}
\usepackage{tcolorbox}
\usepackage{dsfont}
\usepackage{url}
\usepackage[tight]{subfigure}
\usepackage{color}
\usepackage{subfigure}
\usepackage{amsthm}
\usepackage[export]{adjustbox}
\usepackage{comment}
\usepackage[utf8]{inputenc} 
\usepackage[T1]{fontenc}    
\usepackage{url}            
\usepackage{booktabs}       
\usepackage{amsfonts}       
\usepackage{nicefrac}       
\usepackage{microtype}      
\usepackage{diagbox}
\usepackage{balance}
\usepackage{subcaption}
\usepackage{sidecap}
\usepackage{tcolorbox}
\usepackage{pifont}
\usepackage{wrapfig}
\usepackage{longtable} 
\usepackage[linesnumbered,titlenumbered,ruled,vlined,commentsnumbered]{algorithm2e}
\SetKwComment{Comment}{$\triangleright$\ }{}
\definecolor{cvprblue}{rgb}{0.21,0.49,0.74}
\usepackage{siunitx}
\usepackage{mathtools}
\usepackage{soul}
\usepackage{stfloats}
\usepackage{colortbl}
\usepackage{bm}
\usepackage{makecell}
\usepackage{subcaption}
\usepackage{sidecap}
\usepackage{helvet}  
\usepackage{courier}  
\usepackage{babel}
\usepackage{array}

\usepackage[colorlinks,linkcolor=red,citecolor=blue]{hyperref}

\def\argmin{\operatornamewithlimits{arg\,min}}

\usepackage{xspace}
\makeatletter
\DeclareRobustCommand\onedot{\futurelet\@let@token\@onedot}
\def\@onedot{\ifx\@let@token.\else.\null\fi\xspace}
\def\eg{\emph{e.g}\onedot} 
\def\ie{\emph{i.e}\onedot} 
 
\def\etc{\emph{etc}\onedot} 
 
\def\etal{\emph{et al}\onedot}
\makeatother

\makeatletter
\renewcommand{\maketag@@@}[1]{\hbox{\m@th\normalsize\normalfont#1}}%
\makeatother

\ifCLASSOPTIONcompsoc
  \usepackage[nocompress]{cite}
\else
  \usepackage{cite}
\fi
\ifCLASSINFOpdf
\else
\fi
\begin{document}
%
\title{Text Modality Oriented Image Feature Extraction \\ for Detecting Diffusion-based DeepFake}
%
%
%
%

\author{Di~Yang,
        Yihao~Huang,
        Qing~Guo,
        Felix~Juefei-Xu,
        Xiaojun~Jia,
        Run~Wang,
        Geguang~Pu,
        and~Yang~Liu
}
\author{
    \IEEEauthorblockN{Di~Yang$^1$,
        Yihao~Huang$^2$,
        Qing~Guo$^3$,
        Felix~Juefei-Xu$^4$,\\
        Xiaojun~Jia$^2$,
        Run~Wang$^5$,
        Geguang~Pu$^1$,
        and~Yang~Liu$^2$}\\
    \IEEEauthorblockA{$^1$ East China Normal University, China}\\
    \IEEEauthorblockA{$^2$ Nanyang Technological University, Singapore}\\
    \IEEEauthorblockA{$^3$ CFAR and IHPC, Agency for Science, Technology and Research (A*STAR), Singapore}\\
    \IEEEauthorblockA{$^4$ New York University, USA}\\
    \IEEEauthorblockA{$^5$ Wuhan University, China}\\
}

%
%

\markboth{May 2024}%
{Shell \MakeLowercase{\textit{et al.}}: Bare Advanced Demo of IEEEtran.cls for IEEE Computer Society Journals}
%



\IEEEtitleabstractindextext{%
\begin{abstract}
The widespread use of diffusion methods enables the creation of highly realistic images on demand, thereby posing significant risks to the integrity and safety of online information and highlighting the necessity of DeepFake detection. Our analysis of features extracted by traditional image encoders reveals that both low-level and high-level features offer distinct advantages in identifying DeepFake images produced by various diffusion methods. Inspired by this finding, we aim to develop an effective representation that captures both low-level and high-level features to detect diffusion-based DeepFakes. To address the problem, we propose a text modality-oriented feature extraction method, termed TOFE. Specifically, for a given target image, the representation we discovered is a corresponding text embedding that can guide the generation of the target image with a specific text-to-image model. Experiments conducted across ten diffusion types demonstrate the efficacy of our proposed method.
\end{abstract}

}

\maketitle

\IEEEdisplaynontitleabstractindextext

%
\IEEEpeerreviewmaketitle

\ifCLASSOPTIONcompsoc
\IEEEraisesectionheading{\section{Introduction}\label{sec:introduction}}\label{sec:intro}
\else
\section{Introduction}
\label{sec:introduction}\label{sec:intro}
\fi

\IEEEPARstart{W}ith the popularization and development of diffusion technology \cite{croitoru2023diffusion}, recent advancements in the field of synthetic content generation \cite{ho2020denoising,song2020denoising} have marked a new era of generative AI. However, diffusion models can produce highly realistic images that are often indistinguishable for humans from real ones \cite{rombach2022high,zhang2023adding}, potentially disrupting digital media communications and posing new societal risks \cite{riskdeepfake}. Consequently, developing reliable DeepFake detection \cite{juefei2022countering} methods that can keep pace with the latest generative models is extremely challenging.


Regarding DeepFake detection, \textbf{extracting features from real and fake images for further classification is a fundamental and critical step}, as effective and deliberate feature extraction significantly benefits the performance of downstream tasks. However, previous detection methods \cite{wang2023dire,sha2023fake,ricker2024aeroblade,ojha2023towards} generally treat DeepFake detection as a traditional binary classification task similar to image classification, while neglecting the differences between them. Specifically, image classification tasks primarily focus on analyzing the semantics of images, whereas DeepFake detection also requires information on fine-grained details \cite{bayar2018constrained,corvi2023detection}. Trapped in this conventional mindset, most methods predominantly exploit high-level image features (\ie, semantics) while overlooking low-level features (\ie, fine-grained details), which may also be crucial for detection.

Through analyzing the features extracted by the image encoder of two classical model architectures (i.e., CNN \cite{he2016deep} and ViT \cite{dosovitskiy2021an}) on a large number of diffusion-based DeepFakes \cite{dhariwal2021diffusion,ho2020denoising,nichol2021improved,liu2022pseudo,gu2022vector}, we find that low-level features offer advantages in distinguishing real from fake images in some cases, while high-level features are advantageous in other scenarios. Therefore, \textbf{relying solely on either high-level or low-level features to construct a DeepFake detection method is not comprehensive enough.} We suggest finding a representation that properly captures both features.

\begin{figure}
    \centering
    \includegraphics[width=0.9\linewidth]{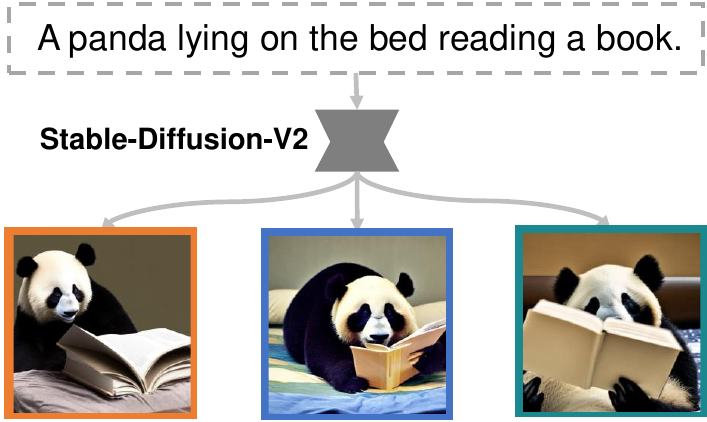}
    \caption{Text to image generation.}
    \label{fig:text_multiple_output_image}
    \vspace{-13pt}
\end{figure}

The idea comes from the observation of the Visual-language model (VLM). In VLM, the \textbf{text (usually represented by discrete tokens) generally reflects high-level semantics and such text has a mutual mapping relationship to the image. This means the text has the potential to act as a starting point for deriving the representation of the image.} However, the text is rough (lacks fine-grained details) and it can be mapped to multiple different images (See Fig.~\ref{fig:text_multiple_output_image}). This requires us to refine such text to contain low-level features. Note that the token text is coarse-grained and discrete, thus in order to adjust text representations more flexibly, we transform it into embedding representation before refining it. In our opinion, once the refined embedding representation can \textit{guide the generation} of the target image in detail through a text-to-image model, it actually captures both high-level and low-level features of the target image.

Inspired by this high-level idea, we propose a novel \textbf{T}ext modality-\textbf{O}riented \textbf{F}eature \textbf{E}xtraction method, termed \textbf{TOFE}. To be specific, given a target real or fake image, TOFE first obtains the embedding of a text input and then iteratively optimizes the embedding to be a representation that can guide the generation of the target image with a pre-trained text-to-image model. The representation shows better performance in distinguishing real and fake images than features extracted from image encoders of classical architecture (\ie, ResNet and CLIP), \textbf{demonstrating the potential of the cross-modal feature extraction method.} 

To summarize, our work has the following contributions:

\begin{itemize}[leftmargin=0.5cm]
\item We are pioneers in suggesting that relying solely on one of the high-level or low-level features to detect diffusion-based DeepFake is not comprehensive enough.
\item To the best of our knowledge, we are the first to explore extracting image features into the text modality for the DeepFake detection task, showing the potential of cross-modal feature extraction.
\item The representation obtained by our TOFE method captures both high-level and low-level features and shows well effectiveness for detecting diffusion-based DeepFake.
\item Experiments conducted on ten diffusion types have verified the effectiveness of our method.
\end{itemize}

\noindent\textbf{Impact and ethical considerations.} In this study, we have solely relied on publicly available data and our main objective is to detect fake information. We firmly assert that the societal benefits stemming from our study far surpass the relatively minor risks of potential harm.


\section{Related Work}
\label{sec:relate}
\subsection{Diffusion-based Fake Image Generation}
In recent years, diffusion models have gradually become the mainstream technology of AI-generated content (AIGC) \cite{zhang2023adding, hertz2023prompttoprompt, ruiz2023dreambooth, van2023anti}. The development was started by Ho \etal{} on proposing a new generation paradigm, denoising diffusion probabilistic models (DDPMs) \cite{ho2020denoising} which achieves competitive performance compared to well-known PGGAN \cite{karras2018progressive}. A lot of works then pay attention to the improvement of architectures, sampling speed, \etc. The denoising diffusion implicit model (DDIM) \cite{song2021denoising} exploits non-Markovian diffusion processes to refine the DDPM for fewer sampling steps and high-fidelity outputs. ADM \cite{dhariwal2021diffusion} proposes a better architecture and achieves image sample quality superior to other generative models. PNDM \cite{liu2022pseudo} further improves sampling efficiency and generation quality with pseudo-numerical methods. 

The previous work focuses on unconditional image synthesis and lacks controllability of the generation procedure. Thus classifier-free guidance diffusion \cite{ho2021classifierfree} explores a more controllable method for image generation, which uses weight to control the balance of fidelity and variety. VQ-Diffusion \cite{gu2022vector} utilizes VQ-VAE \cite{NIPS2017_7a98af17} to handle more complex scenes and improve the synthesized image quality by a large margin. The latent diffusion model (LDM) \cite{rombach2022high} is newly proposed to denoise data in latent space and then decode it to a detailed image, which can significantly improve the generation speed and reduce computing complexity. The famous Stable Diffusion \cite{stable_diffusion} is based on LDM.

\subsection{Detection on Diffusion-based DeepFake}
There is a lot of work \cite{wang2020cnn,frank2020leveraging,yu2019attributing,huang2022fakelocator} pay attention to the detection of GAN-based DeepFake. However, the generation technique of diffusion models and GAN-based generators are entirely different, making previously developed DeepFake detectors ineffective, thus DIRE \cite{wang2023dire} proposes to detect DeepFake by an observation that images generated by diffusion can be reconstructed more accurately with a pre-trained diffusion model than real images. UFD \cite{ojha2023towards} uses the feature space of CLIP for nearest-neighbor classification. It achieves significant performance improvement on unseen diffusion models. AEROBLADE \cite{ricker2024aeroblade} is a training-free detection method for latent diffusion models that uses autoencoder reconstruction error to detect generated images.

The detectors for diffusion-based DeepFake are mostly based on high-level features extracted from CNN or ViT while overlooking the low-level features. In contrast, our work comprehensively evaluates the contribution of high-level and low-level features for distinguishing real and fake images, which calls for future research to integrate them together for better detection performance.

\section{Motivation}\label{sec:motivation}

Features extracted from images are important factors in the DeepFake detection task. Previous detection methods conventionally exploit the high-level features for detection while overlooking the effect of low-level features. Here we conduct feature analysis on the popular DIRE dataset \cite{wang2023dire} which has DeepFake images generated from ten diffusion types. Since CNN-based image encoders are widely used in classification tasks \cite{he2016deep} while features extracted from ViT-based foundation models have surprisingly high performance on the detection task \cite{ojha2023towards}, we choose two classical pre-trained models ResNet50 and CLIP-ViT-L-14 (abbreviate to CLIP) for the empirical study. 

\subsection{Qualitative Analysis}
\begin{figure*}[htb]
\centering
\includegraphics[width=\textwidth]{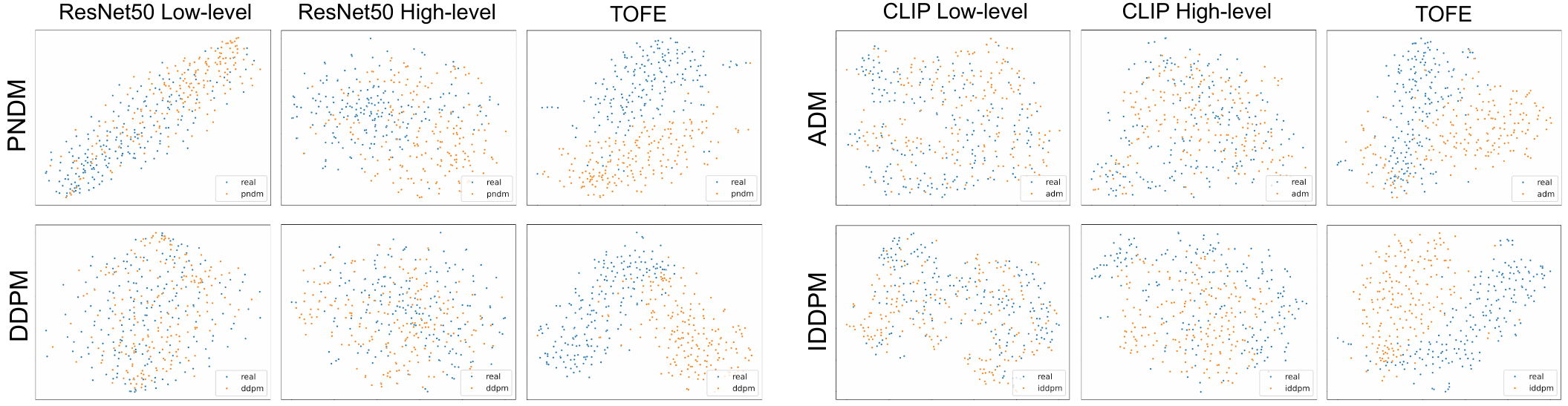}
\caption{T-SNE visualization of features extracted from ResNet, CLIP, and TOFE.}
\label{fig:motivation_tSNE_TOFE}
\end{figure*}
We use t-distributed Stochastic Neighbor Embedding (T-SNE) \cite{vandermaaten08a} to explore and understand the ability of features to distinguish real and fake images. In Figure~\ref{fig:motivation_tSNE_TOFE}, the \textcolor{orange}{orange} and \textcolor{blue}{blue} points represent feature points of real images and fake images respectively. Here we only show the diffusion types that can not be handled by ResNet or CLIP while the complete demonstration of ten diffusion types is in Figure~\ref{fig:TOFE_ResNet_CLIP_feature_TSNE} of Appendix. We can find that the features extracted by ResNet on PNDM \cite{liu2022pseudo} and DDPM \cite{ho2020denoising}, no matter high-level or low-level, are hard to distinguish real images from fake ones (blue and orange sample points are basically intertwined). Also, the features extracted by CLIP on ADM \cite{dhariwal2021diffusion} and IDDPM \cite{nichol2021improved} are not good enough. Compared with them, the representation obtained by our feature extraction method TOFE can distinguish real and fake distributions obviously.

\subsection{Quantitative Analysis}
\begin{table*}[tb]
\center
\caption{Quantitative result of feature extraction methods across different diffusion types.}
\resizebox{\textwidth}{!}{
\begin{tabular}{c|c|c|ccccccccccc}
    \toprule[2pt]
    & \multicolumn{2}{c|}{\textbf{Evaluation}} & \textbf{ADM} & \textbf{DALLE2} & \textbf{DDPM} & \textbf{IDDPM} & \textbf{IF} & \textbf{LDM} & \textbf{PNDM} & \textbf{SD-V1} & \textbf{SD-V2} & \textbf{VQ-Diffusion}\tabularnewline
    \midrule[2pt]
    \multirow{8}{*}{\textbf{ResNet}} & \multirow{2}{*}{\textbf{Low-level}} & \textbf{MMD} $\uparrow$ & \textbf{0.021} & \textbf{1.525} & 0.043 & \textbf{0.355} & \textbf{2.306} & \textbf{1.465} &  0.235 & 0.135 & 0.117 & \textbf{0.310}\tabularnewline
    \cmidrule{3-3} 
    &  & \textbf{JS} $\uparrow$& \textbf{\num{9.479e-5}} & \textbf{\num{1.102e-2}} & \textbf{\num{2.126e-4}} & \textbf{\num{8.388e-3}} & \textbf{\num{2.713e-2}} & \textbf{\num{1.106e-2}} & \num{1.130e-4} & \textbf{\num{1.259e-3}} & \num{2.689e-4} & \num{5.262e-4}\tabularnewline
    \cmidrule{2-13} 
    & \multirow{2}{*}{\textbf{High-level}} & \textbf{MMD} $\uparrow$& 0.006 & 0.520 & \textbf{0.066} & 0.115 & 0.044 & 0.887 &  \textbf{1.144} & \textbf{0.199} & \textbf{1.048} & 0.308\tabularnewline
    \cmidrule{3-3} 
    &  & \textbf{JS} $\uparrow$& \num{2.275e-5} & \num{8.925e-5} & \num{2.948e-5} & \num{1.024e-5} & \num{8.482e-5} & \num{5.092e-5} &  \textbf{\num{7.498e-3}} & \num{1.284e-6} & \textbf{\num{5.671e-3}} & \textbf{\num{7.867e-4}}\tabularnewline
    \cmidrule{2-13}
    & \multirow{2}{*}{\textbf{Low \& High}} & \textbf{MMD} $\uparrow$& \underline{0.035} & 0.685 & \underline{0.251} & 0.126 & 1.788 & 1.018 &  0.027 & \underline{3.771} & \underline{2.163} & \underline{2.123}\tabularnewline
    \cmidrule{3-3} 
    &  & \textbf{JS} $\uparrow$& \num{1.066e-5} & \num{1.017e-3} & \underline{\num{1.212e-3}} & \num{7.265e-4} & \num{8.513e-3} & \num{4.203e-4} &  \num{9.867e-5} & \underline{\num{1.826e-2}} & \underline{\num{6.513e-3}} & \underline{\num{1.032e-2}}\tabularnewline
    \midrule[2pt] 
    \multirow{8}{*}{\textbf{CLIP}} & \multirow{2}{*}{\textbf{Low-level}} & \textbf{MMD} $\uparrow$& \textbf{0.104} & 0.749 & \textbf{0.227} & 0.028 & 0.997 & \textbf{0.353} &  0.565 & 0.153 & 0.318 & 0.908\tabularnewline
    \cmidrule{3-3} 
    &  & \textbf{JS} $\uparrow$& \num{2.432e-6} & \num{1.055e-5} & \num{4.584e-6} & \num{2.193e-6} & \num{2.165e-5} & \num{5.747e-6} &  \num{1.705e-5} & \num{6.475e-6} & \num{1.363e-5} & \num{1.145e-5}\tabularnewline
    \cmidrule{2-13} 
    & \multirow{2}{*}{\textbf{High-level}} & \textbf{MMD} $\uparrow$& 0.061 & \textbf{0.836} & 0.057 & \textbf{0.103} & \textbf{3.436} & 0.159 &  \textbf{2.022} & \textbf{3.682} & \textbf{0.697} & \textbf{3.649}\tabularnewline
    \cmidrule{3-3} 
    &  & \textbf{JS} $\uparrow$& \textbf{\num{6.576e-5}} & \textbf{\num{6.427e-4}} & \textbf{\num{1.119e-4}} & \textbf{\num{3.299e-4}} & \textbf{\num{1.806e-2}} & \textbf{\num{1.946e-4}} &  \textbf{\num{9.329e-3}} & \textbf{\num{2.309e-2}} & \textbf{\num{2.615e-3}} & \textbf{\num{2.508e-2}}\tabularnewline
    \cmidrule{2-13}
    & \multirow{2}{*}{\textbf{Low \& High}} & \textbf{MMD} $\uparrow$& 0.019 & \underline{5.145} & \underline{0.931} & \underline{0.456} & 2.105 & \underline{2.245} &  1.398 & \underline{4.463} & \underline{5.753} & 1.274\tabularnewline
    \cmidrule{3-3} 
    &  & \textbf{JS} $\uparrow$& \num{3.801e-6} & \underline{\num{3.409e-2}} & \underline{\num{3.878e-3}} & \underline{\num{1.841e-3}} & \num{1.001e-2} & \underline{\num{7.706e-3}} &  \num{5.464e-4} & \underline{\num{2.969e-2}} & \underline{\num{4.328e-2}} & \num{4.339e-3}\tabularnewline
    \midrule[2pt] 
    \multicolumn{2}{c|}{\multirow{2}{*}{\textbf{TOFE (ours)}}} & \textbf{MMD} $\uparrow$& \cellcolor{blue!15}{1.569} & \cellcolor{blue!15}{5.373} & \cellcolor{blue!15}{2.931} & \cellcolor{blue!15}{2.729} & 3.155 & \cellcolor{blue!15}{5.825} &  \cellcolor{blue!15}{2.996} & \cellcolor{blue!15}{5.123} & 5.078 & 2.493\tabularnewline
    \cmidrule{3-3} 
        \multicolumn{2}{c|}{} & \textbf{JS} $\uparrow$& \cellcolor{blue!15}{\num{1.159e-2}} & \cellcolor{blue!15}{\num{1.167e-1}} & \cellcolor{blue!15}{\num{4.011e-2}} & \cellcolor{blue!15}{\num{2.464e-2}} & \num{1.393e-2} & \cellcolor{blue!15}{\num{1.260e-1}} &  \cellcolor{blue!15}{\num{2.459e-2}} & \cellcolor{blue!15}{\num{1.100e-1}} & \cellcolor{blue!15}{\num{9.244e-2}} & \cellcolor{blue!15}{\num{2.521e-2}}\tabularnewline
    \bottomrule[2pt]
    \end{tabular}
    }
\label{tab:empirial_feature_extraction}
\end{table*}
 
Only demonstrating qualitative results may not be clear and objective, thus we show the quantitative results on features extracted by ResNet, CLIP, and TOFE.
In Table~\ref{tab:empirial_feature_extraction}, the features are processed by T-SNE and then we use quantitative metrics such as Maximum Mean Discrepancy (MMD) \cite{gretton2012kernel}, Jensen–Shannon divergence (JS) \cite{lin1991divergence} to evaluate the distance between real point distribution and fake point distribution (\ie, the distance between distributions of orange and blue points in Figure~\ref{fig:motivation_tSNE_TOFE}). Regarding these two metrics, the larger the value, the farther the real-fake distribution, and the better the extracted features. For the ``ResNet'' and ``CLIP'' rows, we \textbf{bold} the higher value between low-level and high-level features. Regarding the ``ResNet'' row, we can find that the low-level feature is a bit better at distinguishing real and fake images. Regarding the ``CLIP'' row, the high-level feature is significantly better than low-level ones but not on all diffusion types. The observation inspires us that, \textbf{low-level and high-level features are not absolutely better than each other, it is not comprehensive enough to build a DeepFake detection method just relying on a single of them.}

A naive idea is to directly concatenate the high-level and low-level features together as the representation for detection. In Table~\ref{tab:empirial_feature_extraction}, we also calculate the quantitative results of fused low-level and high-level features in ``Low \& High'' rows. Regarding the ``ResNet'' and ``CLIP'' rows, we respectively \underline{underline} the values which are both higher than that in low-level and high-level features. We can find that, the fused feature shows better performance in some cases, \textbf{indicating that a representation capturing both high-level and low-level features has the potential to benefit the task but simply concatenating them is not good enough.} Thus we not only have to fuse features, but also the fused representation needs to benefit the detection task, which is a challenge.

In the last row, we show the quantitative results of features extracted by our method TOFE. We use \colorbox{blue!15}{blue} cell to label the values which are higher than all the values in the same column (\ie, same diffusion type). We can find that the features extracted by TOFE show obvious advantages in distinguishing real and fake images, which makes features easier for the classifier to learn.

\section{Text Modality Oriented Feature Extraction Method}\label{sec:method}
Inspired by the fact that the text of the text-to-image (T2I) model is a high-level semantic representation, our text modality-oriented feature extraction method aims to obtain the representation by refining the embedding of text to contain fine-grained details (\ie, low-level features). \textit{Note that although the text of the text-to-image model is a high-level semantic representation, this is because it is made up of discrete tokens. The embedding of the text is in continuous space and can be moved continuously in the embedding space (with optimization) to a target embedding representation that contains high-level and low-level information.}

\subsection{Preliminary}
Our feature extraction method is over the Latent Diffusion Model (LDM) \cite{rombach2022high}, a variant of the Denoising Diffusion Probabilistic Model (DDPM) \cite{ho2020denoising} that operates in the latent space of an autoencoder.

\noindent\textbf{Latent diffusion model.} 
The conditional text-to-image LDM is designed to map a noise vector $\mathbf{z}_T$ and text condition $\mathcal{Q}$ to an output latent vector $\mathbf{z}_0$. In order to perform sequential denoising, the network $\epsilon_{\theta}(\cdot)$ is trained to predict the noise at each timestep, following the objective:
\begin{align}
\mathop{\min}_{\theta} \mathbb{E}_{\mathbf{z}, \epsilon \sim \mathcal{N}(0,1),t \sim \mathrm{Uniform}(1,T)} \|\epsilon - \epsilon_\theta(\mathbf{z}_t, t, \mathcal{C})\|^2_2, 
\label{eq:LDM_objective}
\end{align}
where $\mathcal{C} = \Psi(\mathcal{Q})$ is the embedding of the text condition $\mathcal{Q}$, symbol $T$ is the total time steps in the sequential denoising procedure, and $\mathbf{z}_t$ is a noise sample at timestep $t$. When denoising the latent, given the noise vector $\mathbf{z}_T$, the noises sequentially predicted by $\epsilon_{\theta}(\cdot)$ can be gradually removed for $T$ steps with DDIM sampling \cite{song2021denoising},
\begin{scriptsize}
\begin{align}
\mathbf{z}_{t-1} = \sqrt{\frac{\alpha_{t-1}}{\alpha_t}}\mathbf{z}_t + \sqrt{\alpha_{t-1}}\left(\sqrt{\frac{1}{\alpha_{t-1}}-1} -  \sqrt{\frac{1}{\alpha_{t}}-1} \right)\cdot \epsilon_{\theta}(\mathbf{z}_t),
\label{eq:DDIM_sampling}
\end{align}
\end{scriptsize}
where the definition of $\alpha_{t}$ can refer to DDIM.

\noindent\textbf{Classifier-free guidance.}
For conditional text-to-image generation, Ho \etal{} \cite{ho2021classifierfree} propose the classifier-free guidance technique which fuses the predictions performed conditionally and unconditionally to guide the sampling procedure, which can generate arbitrary image categories. To be specific, let $\oslash = \Psi("")$ to be the embedding of a null text, the prediction is defined by 
\begin{align}
\hat{\epsilon_\theta}(\mathbf{z}_t, t, \mathcal{C}, \oslash) = w \ast \epsilon_{\theta}(\mathbf{z}_t, t, \mathcal{C}) + (1-w) \ast \epsilon_{\theta}(\mathbf{z}_t, t, \oslash), 
\label{eq:classifier_free_prediction}
\end{align}
where $w$ is the guidance scale parameter and $w=7.5$ is default in Stable Diffusion \cite{rombach2022high} (\ie, a popular and well-known LDM variant).

\begin{figure}[tb]
\centering
\includegraphics[width=\linewidth]{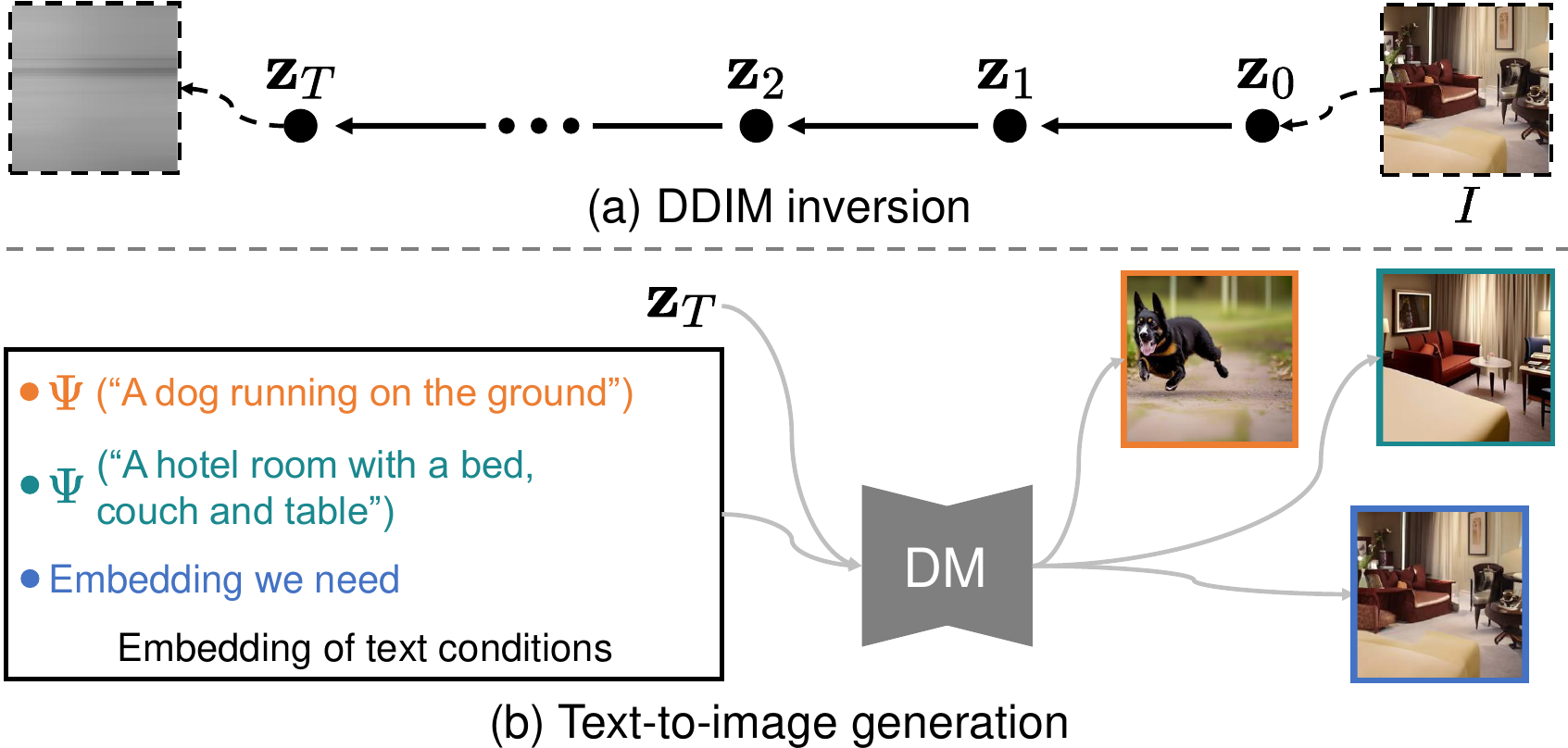}
\caption{Text-to-image generation with different conditions.}
\label{fig:method_motivation}
\end{figure}

\noindent\textbf{DDIM inversion.}
In contrast to DDIM sampling, DDIM \cite{song2021denoising} also propose a simple inversion technique which can gradually add noise to $\mathbf{z}_0$ for $T$ timesteps to achieve $\mathbf{z}_T$ (see Figure~\ref{fig:method_motivation}(a)). The method is based on the assumption that the ordinary differential equation (ODE) process can be reversed in limited small steps that:
\begin{scriptsize}
\begin{align}
\mathbf{z}_{t+1} = \sqrt{\frac{\alpha_{t+1}}{\alpha_t}}\mathbf{z}_t + \sqrt{\alpha_{t+1}}\left(\sqrt{\frac{1}{\alpha_{t+1}}-1} -  \sqrt{\frac{1}{\alpha_{t}}-1} \right)\cdot \epsilon_{\theta}(\mathbf{z}_t). 
\label{eq:DDIM_inversion}
\end{align}
\end{scriptsize}

\subsection{Problem Formulation and Solution}
\noindent\textbf{Observation and motivation.} As shown in Figure~\ref{fig:method_motivation}(a), given a vector $\mathbf{z}_0$ which is latent of an image $I$, we can use DDIM inversion to reverse $T$ timesteps to achieve the latent $\mathbf{z}_T$ (in this example, $T=50$). With the $\mathbf{z}_T$, as shown in Figure~\ref{fig:method_motivation}(b), using classifier-free guidance to generate images with different text conditions $\mathcal{Q}$ will lead to distinct output latents which represent various image contents. There are two observations. \ding{182} Regarding embedding of the first (\textcolor{orange}{orange}) text condition, due to its completely inconsistent high-level information from the image $I$, the generated image (image with \textcolor{orange}{orange} border) is significantly different from image $I$. \ding{183} Regarding embedding of second (\textcolor{teal}{teal}) text condition, it is extracted from $I$ by BLIP \cite{li2022blip}, which shares consistent high-level semantic information with image $I$. However, due to a lack of description of fine-grained details in the embedding, relying solely on high-level semantic information can only result in an image (image with \textcolor{teal}{teal} border) that differs greatly in details from the image $I$. From the previous observations, we can find that, \textbf{in order to reconstruct the image $I$, we need an embedding (\textcolor{blue}{blue}) that contains high-level and low-level features together and it is the representation that we need for further detection tasks.} The problem is how to obtain such embedding representation.

\noindent\textbf{Problem definition.}
Given a target real or fake image $I$ (corresponding latent is $\mathbf{z}_0$) and a pre-trained conditional text-to-image LDM $\mathbf{DM}(\cdot)$, the latent trajectory $\mathbf{z}_1$, $\dots$, $\mathbf{z}_T$ is achieved by DDIM inversion operation with $t = 1, \dots, T$ timesteps respectively. In order to reconstruct $\mathbf{z}_0$, the procedure should start with latent $\mathbf{z}_T$ and perform classifier-free guidance generation with the same condition $\mathcal{C}_t$ (embedding of a token text with high-level semantic) at each timestep $t$ to follow the reverse latent trajectory (\ie, $\mathbf{z}_T$, $\mathbf{z}_{T-1}$, $\dots$, $\mathbf{z}_0$). For each timestep $t$, due to the coarse-grained description of $\mathcal{C}_t$, there is a deviation between the generated latent $\mathbf{z}^{\ast}_{t-1}$ and ground truth latent $\mathbf{z}_{t-1}$ from trajectory. \textbf{Our goal} is to obtain $\hat{\mathcal{C}}_t$ that can make the $\mathbf{z}^{\ast}_{t-1}$ to be same as $\mathbf{z}_{t-1}$ by refining the condition $\mathcal{C}_t$. The $\hat{\mathcal{C}}_t$ is a representation that can guide the generation of the target image, which means it captures the high-level and low-level information of the target image (satisfying our requirement). 



\begin{figure}[tb]
\centering
\includegraphics[width=\linewidth]{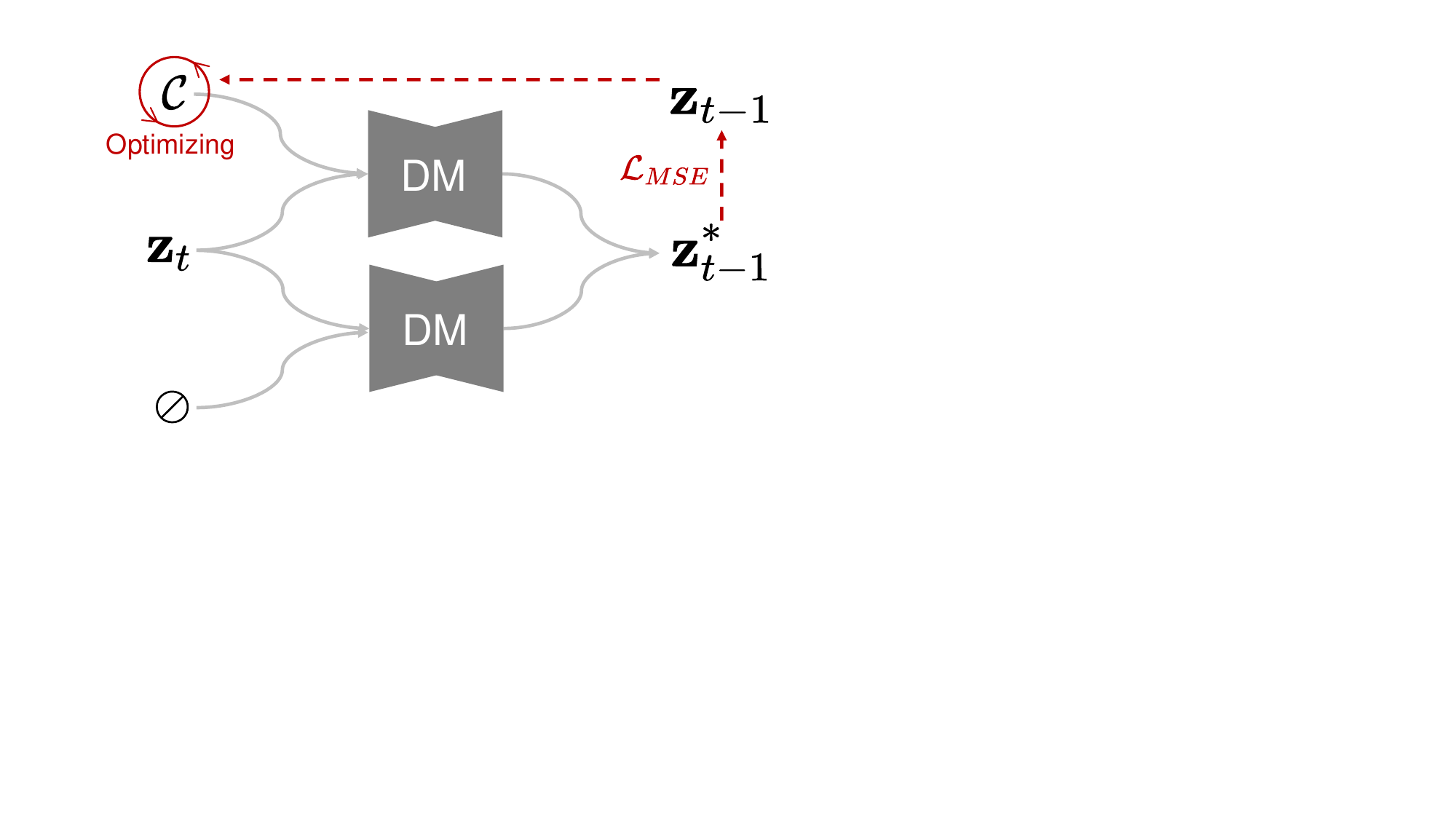}
\caption{Overview of optimization on condition embedding.}
\label{fig:optimizing_condition}
\end{figure}

\noindent\textbf{Objective.} To solve the problem, the idea is to start from a condition embedding $\mathcal{C}$  (\ie, $\mathcal{C} = \Psi(\mathcal{Q})$) of a token text condition $\mathcal{Q}$ (\eg, $\mathcal{Q}$=``a dog'') and iteratively optimize it to be $\hat{\mathcal{C}}$. Specifically, as shown in Fig.~\ref{fig:optimizing_condition}, 
%
%
by calculating the mean square error between $\mathbf{z}_{t-1}$ and $\mathbf{z}_{t-1}^{\ast}$ as loss, the objective is
\begin{align}
\hat{\mathcal{C}} = \mathop{\argmin}_{\mathcal{C}} \|\mathbf{z}_{t-1} - \mathbf{z}_{t-1}^{\ast}\|^2_2. 
\label{eq:loss_objective}
\end{align}
Then, for the full timesteps $t = T, \dots, 1$, we optimize $N$ iterations for each condition $\mathcal{C}_t$ and totally achieve $T$ condition embeddings. Each optimized condition embedding $\hat{\mathcal{C}}_t$ captures high-level and low-level features, which is the representation we need for further detection tasks. The full feature extraction method is shown in Algorithm~\ref{alg:tofe_algo}. With the extracted features, we can easily construct a detection model by training a simple classifier (\ie, MLP). Note that although the proposed TOFE can obtain representation capturing both high-level and low-level features while is effective for the DeepFake detection task. \textit{It does not imply that TOFE is the best text modality-oriented feature extraction method and this does not mean any representation that contains high-level \& low-level features can benefit the detection task. Additionally, while TOFE performs well for the detection task, this does not preclude the feasibility of other methods, possibly image modality-oriented feature extraction methods.}

\begin{algorithm}
	{
		\caption{Text modality Oriented Feature Extraction (TOFE)}\label{alg:tofe_algo}
		\KwIn{Condition embeddings $\{\mathcal{C}_t\}^{T}_{t=1}$, Latent $\mathbf{z}_0$ from $I$, Learning rate $\eta$, Iteration number $N$}
		\KwOut{Optimized condition embeddings $\{\hat{\mathcal{C}_t}\}^{T}_{t=1}$}
        \textcolor{blue}{$\rhd$ compute the intermediate results with DDIM inversion over latent $\mathbf{z}_0$}\\
        $\{\mathbf{z}_t\}^{T}_{t=0} \gets \mathbf{Inversion}(\mathbf{z}_0)$  \label{line:ddim_inversion}\\
        
        \textcolor{blue}{$\rhd$ iteratively optimize condition embeddings for each timestep}\\
        \For{$t = T, T-1, \dots, 1$}
        {
            \For{$i = 0, 1, \dots, N-1$}
            {
                $\mathcal{C}_t \gets \mathcal{C}_t - \eta \bigtriangledown_{\mathcal{C}} \|\mathbf{z}_{t-1} - \mathbf{z}_{t-1}^{\ast}\|^2_2 $
            }
        }
        \Return $\{\hat{\mathcal{C}_t}\}^{T}_{t=1}$
	}
\end{algorithm}

\section{Experiment}\label{sec:experiment}
\subsection{Experimental Setups}
\noindent\textbf{Datasets and models.} In our evaluations, we use the DIRE dataset. It contains ten diffusion types, \ie, ADM \cite{dhariwal2021diffusion}, DDPM \cite{ho2020denoising}, IDDPM \cite{nichol2021improved}, DALLE2 \cite{ramesh2022hierarchical}, IF \cite{IF_model}, PNDM \cite{liu2022pseudo}, VQ-Diffusion \cite{gu2022vector}, LDM \cite{rombach2022high}, Stable Diffusion v1 and v2. For each type, there are thousands of fake images. The images are all with size 256 $\times$ 256. The baselines are 
DIRE \cite{wang2023dire}, UFD \cite{ojha2023towards}, AEROBLADE \cite{ricker2024aeroblade}. They are all published and are state-of-the-art detection works for diffusion-based DeepFake. 

\noindent\textbf{Metrics.} We use average accuracy (ACC) and average precision (AP) as the metrics for evaluating the detection performance. The threshold for computing accuracy is set to 0.5 following \cite{wang2023dire}.

\noindent\textbf{Implementation details of our method.} The LDM we used is Stable Diffusion v2.0. For the learning rate $\eta$ and iteration number $N$, the value is 0.01 and 10. The default condition embedding $\mathcal{C}$ is $\oslash = \Psi("")$. Using $\oslash$ is general for any image and resource-consuming (does not need to generate a caption for each image with BLIP). The timestep $T$ is 1 and the reason is that the time consumption of feature extraction is faster than that of other values and the extracted feature is good for detecting real and fake images. Note that when $T$ is other values (\eg, 50), the features are also good enough, as shown in Appendix Section~\ref{sec:TOFE_timestep_50}. The interconversion of images and corresponding latents are based on the autoencoder of the pre-trained stable diffusion model. The self-built classifier for feature analysis is a simple MLP that only contains four layers and cross-entropy loss. The learning rate for the classifier is set to \num{1e-5} and the classifier is trained for 10,000 iterations. All the experiments were run on an Ubuntu system with two NVIDIA A6000 Tensor Core GPUs of 48G RAM.

\subsection{Comparison with DeepFake Detection Baseline}
\begin{table*}[htb]
\center
\setlength{\tabcolsep}{2.5pt}
\caption{Detection performance comparison with state-of-the-art detection methods.}
\resizebox{\linewidth}{!}{
\begin{tabular}{cc|cccccccccc|c}
\toprule[2pt] 
\multirow{2}{*}{} & \multirow{2}{*}{ACC(\%)/AP(\%)} & \multicolumn{10}{c|}{Diffusion Types} & \multirow{2}{*}{Average}\tabularnewline
 &  & ADM \textsuperscript{*} & PNDM \textsuperscript{*} & IDDPM \textsuperscript{*}& DDPM & IF & LDM & DALLE2 & SD-V1 & SD-V2 & VQ-Diffusion & \\ \midrule[2pt]
& UFD \cite{ojha2023towards} & 90.30/97.81 & 98.30/99.95 &98.15/99.87 & 98.50/99.90 &54.50/74.32 & 65.95/84.98 & 95.10/99.01 & 83.15/95.74 & 79.80/93.74 & 98.25/99.89 & 86.20/94.52\tabularnewline
& DIRE \cite{wang2023dire} & 87.60/99.59 & 94.60/99.87 & 90.35/99.67 & 92.42/99.54 & 98.03/98.92 & 87.95/99.51 & 98.15/99.91 & 96.60/99.90 & 98.60/99.97 & 91.65/99.79 & 93.60/\textbf{99.67}\tabularnewline
& AEROBLADE \cite{ricker2024aeroblade} & -/74.26 & -/49.77 & -/60.10 & -/53.66 & -/97.02 & -/99.72 & -/67.48 & -/79.63 & -/87.94 & -/88.88 & -/75.85\\ \midrule
\multicolumn{2}{c|}{TOFE (ours)} & 97.70/99.72 & 98.70/99.90 & 98.75/99.95 & 98.40/99.66 & 97.35/99.73 & 98.25/99.77 & 97.50/99.37 & 96.20/99.50 & 93.45/98.94 & 98.45/99.94 & \textbf{97.48}/99.65\tabularnewline
\bottomrule[2pt]
\end{tabular}}
\label{tab:compare_with_other_baseline}
\end{table*}
As a DeepFake detection method, we compare our TOFE method with other state-of-the-art detection methods, \ie, UFD \cite{ojha2023towards}, DIRE \cite{wang2023dire}, and AEROBLADE \cite{ricker2024aeroblade}. To verify the in-domain (ID) and out-of-domain (OOD) performance of our method, we follow the common practices ``train-on-many and test-on-many'' in DeepFake detection \cite{wang2023dire}, that is, train on some diffusion types and test on more. To be specific, following DIRE \cite{wang2023dire}, we train our classifier on 30,000 images generated by three types of diffusion models (\ie, ADM, PNDM, and IDDPM, with $\textsuperscript{*}$ in Table) and 30,000 real images, a total of 60,000 images. For the test dataset, there are 10 different diffusion types (\ie, ADM, DALLE2, DDPM, IDDPM, IF, LDM, PNDM, SD-V1, SD-V2, and VQ-Diffusion). For each diffusion type, there are 1,000 real images and 1,000 fake images, a total of 2,000 images per type. The baselines (UFD, DIRE, and AEROBLADE) are all trained and tested on the above dataset with their own experimental setting. Since AEROBLADE does not provide a specific way to calculate ACC based on their special setting, we just provide the AP value as they do.

As shown in Table~\ref{tab:compare_with_other_baseline}, we report ACC (\%) and AP (\%) (ACC/AP in the Table). We can find that the detection results of our method achieve the best ACC and second-best AP on average. Since the AP result of our method is 99.65\% and the best AP achieved by ``DIRE'' is just 99.67\% (both very close to 100\%), we think our result is a bit better by taking both ACC and AP into consideration. For in-domain testing (\ie, results on ADM, PNDM, and IDDPM), UFD, DIRE, and our method TOFE all show good performance. For OOD testing, our method and DIRE achieve the best performance, \ie, ACC higher than 90\% and AP higher than 99\% in most cases. To summarize, our method is comparable to DIRE and outperforms UFD and AEROBLADE.

\subsection{Comparison with Other Feature Extraction Methods}
\begin{table*}[tb]
\center
\setlength{\tabcolsep}{2.5pt}
\caption{Comparison with classical feature extraction methods (all with same classifier architecture).}
\resizebox{\linewidth}{!}{
\begin{tabular}{cc|cccccccccc|c}
\toprule[2pt] 
\multirow{2}{*}{} & \multirow{2}{*}{ACC(\%)/AP(\%)} & \multicolumn{10}{c|}{Diffusion Types} & \multirow{2}{*}{Average}\tabularnewline
 &  & ADM\textsuperscript{*} & PNDM\textsuperscript{*} & IDDPM\textsuperscript{*} & DDPM & IF & LDM & DALLE2 & SD-V1 & SD-V2 & VQ-Diffusion & \\ \midrule[2pt]
\multirow{3}{*}{\rotatebox{90}{ResNet}} & Low-level & 72.45/79.26 & 72.45/76.58 & 76.95/82.04 & 73.20/64.56 & 62.60/65.05 & 70.65/73.82 & 54.35/38.71 & 59.35/61.03 & 65.20/69.38 & 68.90/73.42 & 67.61/68.38\tabularnewline
 & High-level & 92.15/97.83 & 96.10/99.44 & 96.15/99.46 & 94.80/97.90 & 93.80/84.30 & 84.40/94.06 & 48.45/27.99 & 47.30/41.00 & 47.45/38.69 & 60.25/76.70 & 76.09/75.74\tabularnewline
 & Low \& High & 94.09/97.99 & 98.88/99.47 & 98.78/99.48 & 97.51/98.14 & 64.01/84.53 & 85.63/94.88 & 15.88/28.17 & 20.33/39.96 & 47.95/94.20 & 55.70/79.80 & 67.88/81.66\\ \midrule
\multirow{3}{*}{\rotatebox{90}{CLIP}} & Low-level & 81.95/91.31 & 91.15/97.87 & 87.20/95.33 & 89.75/93.90 & 70.95/81.71 & 84.80/92.79 & 85.05/87.05 & 77.15/86.16 & 77.80/86.89 & 90.15/96.84 & 83.60/90.98\tabularnewline
 & High-level & 97.85/99.81 & 99.70/99.99 & 99.45/99.98 & 99.15/99.93 & 87.80/98.61 & 83.10/97.97 & 94.80/99.31 & 94.90/99.66 & 94.50/99.60 & 99.45/99.98 & 95.07/99.49\tabularnewline
 & Low \& High & 97.80/99.85 & 99.80/100.00 & 99.60/99.99 & 99.35/99.94 & 87.40/98.80 & 92.15/99.52 & 93.80/99.23 & 95.85/99.75 & 97.05/99.80 & 99.70/100.00 & 96.25/\textbf{99.69}\\ \midrule
\multicolumn{2}{c|}{TOFE (ours)} & 97.70/99.72 & 98.70/99.90 & 98.75/99.95 & 98.40/99.66 & 97.35/99.73 & 98.25/99.77 & 97.50/99.37 & 96.20/99.50 & 93.45/98.94 & 98.45/99.94 & \textbf{97.48}/99.65\tabularnewline
\bottomrule[2pt]
\end{tabular}}
\label{tab:compare_with_other_feature_extraction}
\end{table*}
Although the features extracted by our TOFE method make it easier to distinguish real and fake images than other feature extraction methods, this does not mean that the detection performance of our method is absolutely better since the detection results depend on what the classifier learns. Therefore, we employ the same simple classifier (MLP with only four layers) to process the features from different methods and the detection results are shown in Table~\ref{tab:compare_with_other_feature_extraction}.

We report ACC (\%) and AP (\%) (ACC/AP in the Table). we can find that the detection results of our method achieve the best ACC and second-best AP on average. Since the AP result of our method is 99.65\% and the best AP achieved by ``CLIP Low \& High'' is just 99.69\% (both very close to 100\%), we think our result is a bit better by taking both ACC and AP into consideration. From the table, we can also find two interesting observations. \ding{182} In ``ResNet'' row, the results of ``Low \& High'' is partially better than that of ``Low-level'' and ``High-level'' while in ``CLIP'' row, ``Low \& High'' is absolutely better than that of ``Low-level'' and ``High-level'' on both ACC and AP. This means the effect of simply concatenating features from high-level and low-level for the DeepFake detection task is unstable, which highlights the necessity of finding a good fusion of the features. \ding{183} The detection performance of ``Low-level'', ``High-level'' and ``Low \& High'' in ``CLIP'' row is always better than that in ``ResNet'' row. This provides evidence of the advantages of features extracted from CLIP (foundation model trained with a large dataset), which is consistent with the conclusion in \cite{ojha2023towards}.

\subsection{Ablation Study}
\begin{table*}[tb]
\centering
\caption{Ablation on learning rate $\eta$.}
\resizebox{\linewidth}{!}{
\begin{tabular}{c|c|cccccccccc|c}
\toprule[2pt]
 \textbf{Learning Rate} & \textbf{Metric}  & \textbf{ADM} & \textbf{DALLE2} & \textbf{DDPM} & \textbf{IDDPM} & \textbf{IF} & \textbf{LDM} & \textbf{PNDM} & \textbf{SD-V1} & \textbf{SD-V2} & \textbf{VQ-Diffusion} & \textbf{Average} \\
\midrule[2pt]
\multirow{2}{*}{\textbf{\num{1e-2}}} & \textbf{MMD} $\uparrow$& 1.569 & 5.373 & 2.931 & 2.729 & 3.155 & 5.825 & 2.996 & 5.123 & 5.078 & 2.493 & \textbf{3.727} \\
\cmidrule{2-2} 
 & \textbf{JS} $\uparrow$ & \num{1.159e-2} & \num{1.167e-1} & \num{4.011e-2} & \num{2.464e-2} & \num{1.393e-2} & \num{1.260e-1} & \num{2.459e-2} & \num{1.100e-1} & \num{9.244e-2} & \num{2.521e-2} & \textbf{\num{5.825e-2}} \\
\midrule
\multirow{2}{*}{\textbf{\num{1e-3}}} & \textbf{MMD} $\uparrow$& 0.461 & 1.424 & 2.844 & 0.459 & 0.828 & 3.791 & 3.102 & 5.462 & 5.240 & 0.745 & 2.436 \\
\cmidrule{2-2} 
 & \textbf{JS} $\uparrow$ & \num{5.961e-3} & \num{3.268e-3} & \num{5.009e-2} & \num{4.891e-3} & \num{2.329e-4} & \num{4.662e-2} & \num{3.897e-2} & \num{1.281e-1} & \num{9.679e-2} & \num{2.717e-3} & \num{3.777e-2} \\
\midrule
\multirow{2}{*}{\textbf{\num{1e-4}}} & \textbf{MMD} $\uparrow$& 0.621 & 4.976 & 1.029 & 0.908 & 5.075 & 2.257 & 1.732 & 3.564 & 2.305 & 1.120 & 2.359 \\
\cmidrule{2-2} 
 & \textbf{JS} $\uparrow$ & \num{5.591e-3} & \num{9.280e-2} & \num{4.247e-3} & \num{6.055e-3} & \num{7.583e-2} & \num{3.485e-2} & \num{1.474e-2} & \num{4.877e-2} & \num{1.461e-2} & \num{1.066e-2} & \num{3.082e-2} \\
\midrule
\multirow{2}{*}{\textbf{\num{1e-5}}} & \textbf{MMD} $\uparrow$& 1.527 & 4.679 & 0.853 & 1.140 & 5.379 & 5.739 & 2.643 & 4.413 & 5.713 & 1.490 & 3.358 \\
\cmidrule{2-2} 
 & \textbf{JS} $\uparrow$ & \num{2.091e-2} & \num{7.604e-2} & \num{1.315e-2} & \num{1.111e-2} & \num{8.564e-2} & \num{1.061e-1} & \num{4.811e-2} & \num{7.438e-2} & \num{1.115e-1} & \num{1.009e-2} & \num{5.571e-2} \\
\midrule
\multirow{2}{*}{\textbf{\num{1e-6}}} & \textbf{MMD} $\uparrow$& 0.113 & 1.088 & 0.325 & 0.154 & 1.054 & 1.436 & 1.844 & 0.902 & 1.441 & 1.199 & 0.956 \\
\cmidrule{2-2} 
 & \textbf{JS} $\uparrow$ & \num{3.761e-4} & \num{8.266e-3} & \num{3.959e-3} & \num{8.313e-4} & \num{4.426e-3} & \num{1.799e-2} & \num{2.836e-2} & \num{3.525e-3} & \num{1.265e-2} & \num{9.888e-3} & \num{9.027e-3} \\
\bottomrule[2pt]
\end{tabular}}
\label{tab:ablation_learning_rate}
\end{table*}
We make an ablation study on the two hyperparameters (learning rate $\eta$ and iteration number $N$) to explore their influence on TOFE. The choices of the learning rate are from \num{1e-2} to \num{1e-6}. In Table~\ref{tab:ablation_learning_rate}, following the setting of Table~\ref{tab:empirial_feature_extraction}, we demonstrate the MMD and JS values under different learning rates. We can find that when the learning rate $\eta$ is \num{1e-2}, the MMD and JS value of distributions between real and fake distributions in T-SNE is the highest. Also, we can find that although not the best, the MMD and JS values under \num{1e-3} to \num{1e-5} are comparable or better than that of ResNet and CLIP in Table~\ref{tab:empirial_feature_extraction}. For iteration number, the ablation values are from 10 to 50. We find the changes in the iteration number do not have a significant impact on the results and are all better than that of ResNet and CLIP, thus put the results in Table~\ref{tab:ablation_iteration_number} of Appendix.


\section{Discussion}\label{sec:discussion}
\subsection{Transferability Evaluation on GAN-based DeepFake}
\begin{table}
\center
\setlength{\tabcolsep}{2.5pt}
\caption{Transferability evaluation on GAN-based DeepFake.}
\resizebox{\linewidth}{!}{
\begin{tabular}{cc|ccc|c}
\toprule[2pt] 
\multirow{2}{*}{} & \multirow{2}{*}{ACC(\%)/AP(\%)} & \multicolumn{3}{c|}{GAN Types} & \multirow{2}{*}{Average}
\tabularnewline
 &  & ProGAN & StyleGAN & StarGAN& \\ 
 \midrule[2pt]
& UFD \cite{ojha2023towards} & 97.80/99.74 & 96.70/99.51 &94.16/83.90  &96.22/94.38 \tabularnewline
& DIRE \cite{wang2023dire} & 86.00/99.19 & 91.95/99.74 &83.00/91.48   & 86.98/96.80  \tabularnewline
& AEROBLADE \cite{ricker2024aeroblade} & -/49.33 & -/48.82 & -/43.10 & -/47.08\\ 
\midrule
\multicolumn{2}{c|}{TOFE (ours)} & 97.30/99.66 & 97.95/99.78  &95.30/98.21  &\textbf{96.85}/\textbf{99.22} \tabularnewline
\bottomrule[2pt]
\end{tabular}}
\label{tab:GAN_transferability}
\end{table}
In the image synthesis domain, GAN is another mainstream approach for DeepFake generation. Thus we further explore the transferability of our detection method on the GAN-based DeepFake. Here we choose the classical and well-known types such as ProGAN \cite{karras2018progressive}, StarGAN \cite{choi2018stargan}, and StyleGAN \cite{karras2019style}. For testing on each type, there are 1,000 real and 1,000 fake images. The detection models of UFD, DIRE, AEROBLADE, and our TOFE are the same as in Table~\ref{tab:compare_with_other_baseline}. In Table~\ref{tab:GAN_transferability}, our method achieves the best ACC and AP, which are both higher than 95\%.

\subsection{Reconstruction Quality}
\begin{table*}[tb]
\footnotesize
\caption{Quality of reconstructed images.}
\resizebox{\linewidth}{!}{
\begin{tabular}{c|ccccccccccc}
\toprule
 & Real & ADM & DALLE2 & DDPM & IDDPM & IF & LDM & PNDM & SD-V1 & SD-V2 & VQ-Diffusion\\ \midrule
\textbf{PSNR} & 31.486 & 35.391 & 33.808 & 35.618 & 35.001 & 38.754 & 41.485 & 31.678 & 30.025 & 32.203 & 36.190\tabularnewline
\textbf{SSIM} & 0.891 & 0.935 & 0.911 & 0.942 & 0.927 & 0.970 & 0.977 & 0.871 & 0.874 & 0.873 & 0.943\tabularnewline
\bottomrule
\end{tabular}}
\label{tab:quality_reconstructed_image}
\end{table*}
Since we aim to achieve a representation that can guide the reconstruction of the target image with a specified pre-trained text-to-image model, we evaluate the reconstruction quality to verify whether the representation meets the requirements. Here compared with the target image, we use metrics Peak Signal-to-Noise Ratio (PSNR) \cite{hore2010image} and Structural Similarity (SSIM) \cite{wang2004image} to calculate the quality of the reconstructed image (average of 200 images per type). As shown in Table~\ref{tab:quality_reconstructed_image}, we demonstrate the evaluation result. We can find that the PSNR values are all above 30, which means the reconstructed image and target image are visually identical images \cite{huang2011robust}. For SSIM values, most of the values are larger or near 0.9, which also verifies the high quality of the reconstructed image.

\subsection{Version of T2I model in TOFE}
\begin{table*}[tb]
\center
\setlength{\tabcolsep}{2.5pt}
\caption{Detection performance of TOFE with other Stable Diffusion version.}
\resizebox{\linewidth}{!}{
\begin{tabular}{cc|cccccccccc|c}
\toprule[2pt] 
\multirow{2}{*}{} & \multirow{2}{*}{ACC(\%)/AP(\%)} & \multicolumn{10}{c|}{Diffusion Types} & \multirow{2}{*}{Average}\tabularnewline
 &  & ADM \textsuperscript{*} & PNDM \textsuperscript{*} & IDDPM \textsuperscript{*}& DDPM & IF & LDM & DALLE2 & SD-V1 & SD-V2 & VQ-Diffusion & \\ \midrule[2pt]
\multicolumn{2}{c|}{TOFE (ours)} & 97.35/99.73 & 97.65/99.39 & 98.35/99.60 & 99.05/99.92 & 98.55/99.87 & 98.20/99.81 & 99.15/99.95 & 97.60/99.76 & 96.90/99.59 & 99.30/99.98 & 98.21/99.76\tabularnewline
\bottomrule[2pt]
\end{tabular}}
\label{tab:classifier_Stable_version_1_4}
\end{table*}
We use Stable Diffusion v2.0 as the pre-trained text-to-image model in TOFE for feature extraction in the above experiments. Here we make an ablation study of its version to Stable Diffusion v1.4. The quantitative result and qualitative results are in Appendix Section~\ref{sec:stable_diffusion_version}. We show the detection performance of the classifier (following the setting as in Table~\ref{tab:compare_with_other_baseline}) in Table~\ref{tab:classifier_Stable_version_1_4}. We can find that the performance is even better than all the models in Table~\ref{tab:compare_with_other_baseline}, reflecting that the version of the pre-trained text-to-image model does not affect the result.

\subsection{Robustness Analysis}
\begin{figure}[tb]
\centering
\includegraphics[width=0.9\linewidth]{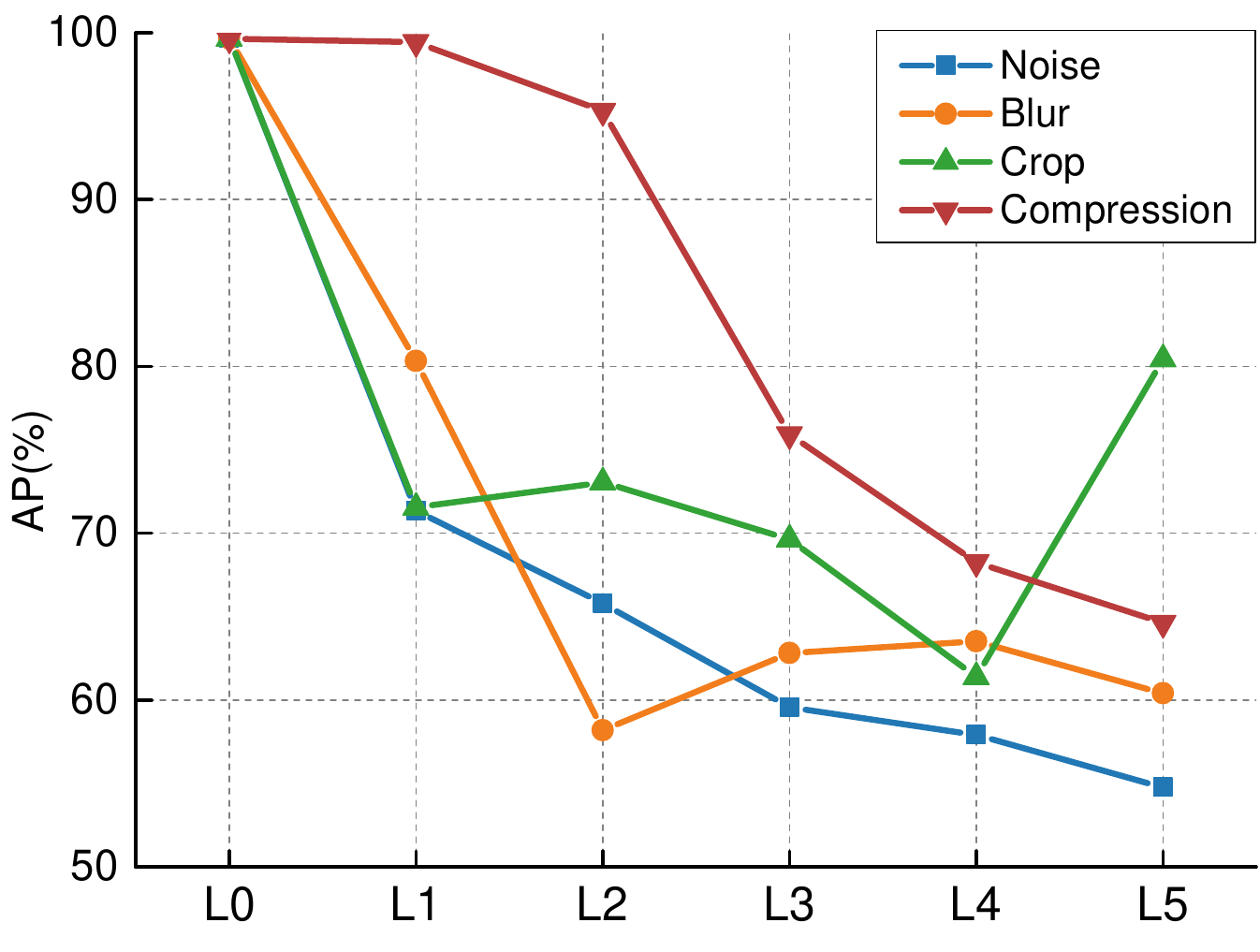}
\caption{Robustness analysis under four corruptions.}
\label{fig:robustness_TOFE}
\end{figure}
In real-world scenarios, images may face different common corruptions when captured by cameras or uploaded to social media. Thus we evaluate how robust the TOFE is against common image corruptions. Referring to \cite{frank2020leveraging,ricker2024aeroblade}, we use Gaussian noise, Gaussian blur, cropping (with subsequent resizing to the original size), and JPEG compression. In Figure~\ref{fig:robustness_TOFE}, we report the AP of TOFE for corruptions with different severities. There are six severities (L0-L5), where L0 is without corruption. For the four corruptions Gaussian noise / Gaussian blur / crop / JPEG compression, following the setting of AEROBLADE \cite{ricker2024aeroblade}, the severity L1-L5 of standard deviation/standard deviation/crop factor/compression quality are 5,10,15,20,25 / 1,2,3,4,5 / 0.9,0.8,0.7,0.6,0.5 / 90,80,70,60,50 (see visualization in Appendix Fig.~\ref{fig:appendix_robustness_image}). The result value is the average across ten diffusion types (types in Table~\ref{tab:compare_with_other_baseline}). We can find that, in general, the AP of the classifier will reduce when corruption severity increases. The AP result under corruption of small severity is acceptable (higher than 70\%) and the AP values are still higher than 50\% under corruption of big severity. Across four corruption types, the classifier has better robustness against JPEG compression than the others. In future work, we plan to add data augmentation ideas into the feature extraction procedure to achieve features that can be trained for a more robust classifier.

\subsection{Limitation}
\begin{table}
\centering
    \caption{Time consumption.}
    \setlength{\tabcolsep}{2.5pt}
    \resizebox{\linewidth}{!}{
    \begin{tabular}{@{}l|cccc|cc@{}}
        \toprule[2pt]
        \textbf{Methods} & \textbf{ResNet Low-level} & \textbf{ResNet High-level} & \textbf{CLIP Low-level} & \textbf{CLIP High-level} & \textbf{TOFE (ours)} \\
        \midrule[2pt]
        \textbf{Time(s)} & 0.043378 & 0.043616 & 0.019569 & 0.019793 & 0.606718 \\
        \bottomrule[2pt]
    \end{tabular}}
    \label{tab:time_comparison}
\end{table}

\textbf{Time consumption.} As an early work, we acknowledge that the proposed feature extraction method still has room for improvement. For example, compared with feature extraction methods by image encoders, our TOFE method has the disadvantage of time consumption. As shown in Table~\ref{tab:time_comparison}, we evaluate the feature extraction procedure with ResNet50, CLIP-ViT-L-14, and TOFE on 1,000 images and calculate the time consumption per image. We can find that CLIP achieves the least time consumption and ResNet is a bit slower than it. Our TOFE method is the slowest and uses 0.6s per image. However, since it is less than 1 second, we think the time consumption is acceptable and aim to reduce the consumption in future work.

\noindent\textbf{Theoretical analysis.} 
Although we propose an effective feature extraction method for DeepFake detection, it would be better to give a theoretical analysis of why it is useful. Since what kind of clues can distinguish between real and fake images remains to be explored, it is hard for us to give further explanation. However, we firmly believe that our exploration is essential and serves as a valuable starting point for cross-modal feature extraction research in the DeepFake detection task.

\section{Conclusion}
With the booming development of diffusion technology, it is more and more difficult to distinguish real images from fake images. In this paper, we propose to extract image features into text modality to achieve a representation that captures both high-level and low-level features. The classifier based on such representation shows amazing performance in detecting diffusion-based DeepFake. In the future, we aim to reduce the time consumption of the feature extraction method.

%
%
%
%


%

\ifCLASSOPTIONcaptionsoff
  \newpage
\fi



%

\bibliographystyle{IEEEtran}
\bibliography{sample-base}

\onecolumn
\newpage
\appendix

\section{Appendix / supplemental material}
\subsection{T-SNE Visualization of Various Feature Extraction Methods}
\begin{figure*}[htb]
    \centering
    \includegraphics[height=0.85\textheight]{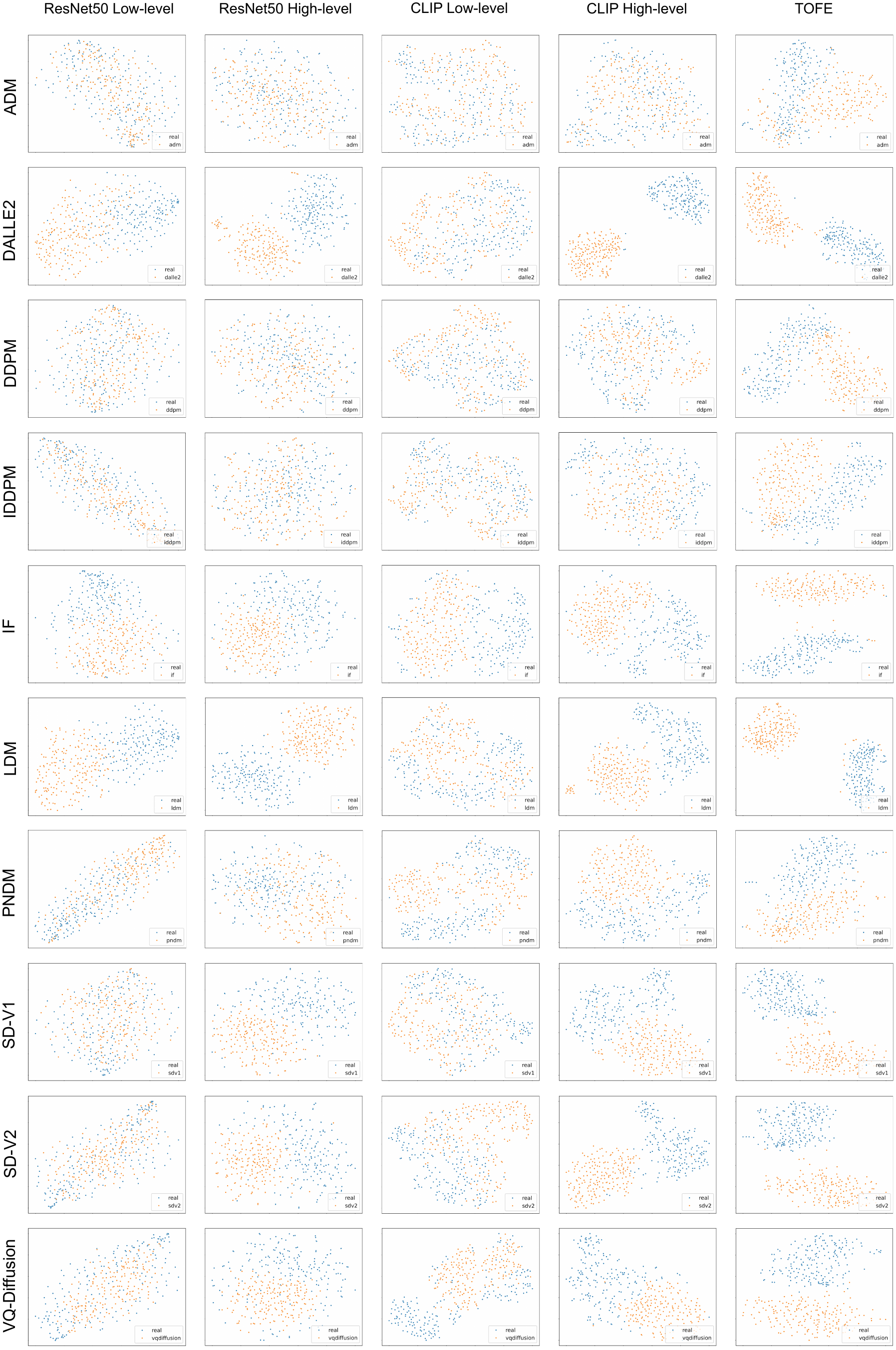}
    \caption{T-SNE visualization of various feature extraction methods across different diffusion types.}
    \label{fig:TOFE_ResNet_CLIP_feature_TSNE}
\end{figure*}

\subsection{Details of Low level and High level in ResNet and CLIP}
For ResNet50, we extract the output of ``layer1'' (dimension is [1,256,56,56]) as the low-level feature and extract the output of ``average pooling'' (prior layer of the linear layer, after ``layer4'', dimension is [1,2048,1,1]) as the high-level feature. For CLIP-ViT-L-14, we extract the output of visual transformer ``resblocks0'' (dimension is [257,1,1024]) as the low-level feature and extract the output of ``layernorm'' (prior layer of the linear layer, after transformer ``resblocks11'', dimension is [1,768]) as the high-level feature.

\subsection{Ablation on Iteration Number}
Here we demonstrate the ablation study on iteration number. In Table~\ref{tab:ablation_iteration_number}, we can find that with different iteration numbers, the average MMD and JS have small differences and are \textbf{all better than} those of ResNet and CLIP.
\begin{table}[htb]
\centering
\caption{Ablation on iteration number $N$.}
\resizebox{\linewidth}{!}{
\begin{tabular}{c|c|cccccccccc|c}
\toprule[2pt]
\textbf{Iteration Number} & \textbf{Metric} & \textbf{ADM} & \textbf{DALLE2} & \textbf{DDPM} & \textbf{IDDPM} & \textbf{IF} & \textbf{LDM} & \textbf{PNDM} & \textbf{SD-V1} & \textbf{SD-V2} & \textbf{VQ-Diffusion} & \textbf{Average} \\
\midrule[2pt]
\multirow{2}{*}{\textbf{10}} & \textbf{MMD} $\uparrow$ & 1.569 & 5.373 & 2.931 & 2.729 & 3.155 & 5.825 & 2.996 & 5.123 & 5.078 & 2.493 & \textbf{3.727} \\
\cmidrule{2-2} 
 & \textbf{JS} $\uparrow$ & \num{1.159e-2} & \num{1.167e-1} & \num{4.011e-2} & \num{2.464e-2} & \num{1.393e-2} & \num{1.260e-1} & \num{2.459e-2} & \num{1.100e-1} & \num{9.244e-2} & \num{2.521e-2} & \num{5.825e-2} \\
\midrule
\multirow{2}{*}{\textbf{20}} & \textbf{MMD} $\uparrow$ & 0.321 & 5.887 & 2.689 & 0.982 & 2.657 & 6.139 & 1.919 & 4.868 & 5.469 & 1.442 & 3.237 \\
\cmidrule{2-2} 
 & \textbf{JS} $\uparrow$ & \num{1.833e-3} & \num{1.383e-1} & \num{3.128e-2} & \num{1.322e-2} & \num{4.798e-2} & \num{1.486e-1} & \num{1.182e-2} & \num{8.151e-2} & \num{1.220e-1} & \num{3.333e-3} & \num{5.999e-2} \\
\midrule
\multirow{2}{*}{\textbf{30}} & \textbf{MMD} $\uparrow$ & 1.910 & 5.750 & 0.510 & 1.809 & 1.679 & 4.948 & 1.130 & 5.529 & 6.011 & 2.091 & 3.137 \\
\cmidrule{2-2} 
 & \textbf{JS} $\uparrow$ & \num{1.567e-2} & \num{1.334e-1} & \num{6.156e-3} & \num{2.270e-2} & \num{1.105e-2} & \num{9.560e-2} & \num{7.487e-3} & \num{1.187e-1} & \num{1.478e-1} & \num{1.350e-2} & \num{5.720e-2} \\
\midrule
\multirow{2}{*}{\textbf{40}} & \textbf{MMD} $\uparrow$ & 1.213 & 5.834 & 0.089 & 0.131 & 2.269 & 6.304 & 0.732 & 4.108 & 4.674 & 1.038 & 2.639 \\
\cmidrule{2-2} 
 & \textbf{JS}$\uparrow$ & \num{1.176e-2} & \num{1.386e-1} & \num{1.287e-4} & \num{1.395e-4} & \num{1.100e-2} & \num{1.473e-1} & \num{9.331e-4} & \num{7.888e-2} & \num{7.943e-2} & \num{1.391e-4} & \num{4.684e-2} \\
\midrule
\multirow{2}{*}{\textbf{50}} & \textbf{MMD} $\uparrow$ & 2.524 & 5.729 & 2.165 & 0.972 & 3.973 & 6.264 & 1.467 & 4.630 & 5.258 & 1.008 & 3.399 \\
\cmidrule{2-2} 
 & \textbf{JS} $\uparrow$ & \num{2.509e-2} & \num{1.365e-1} & \num{2.513e-3} & \num{1.354e-2} & \num{7.216e-2} & \num{1.386e-1} & \num{9.320e-3} & \num{9.080e-2} & \num{1.014e-1} & \num{3.977e-4} & \textbf{\num{6.130e-2}} \\
\bottomrule[2pt]
\end{tabular}}
\label{tab:ablation_iteration_number}
\end{table}

\subsection{Ablation on Version of Stable Diffusion }\label{sec:stable_diffusion_version}
\begin{figure*}[htb]
    \centering
    \includegraphics[width=0.9\linewidth]{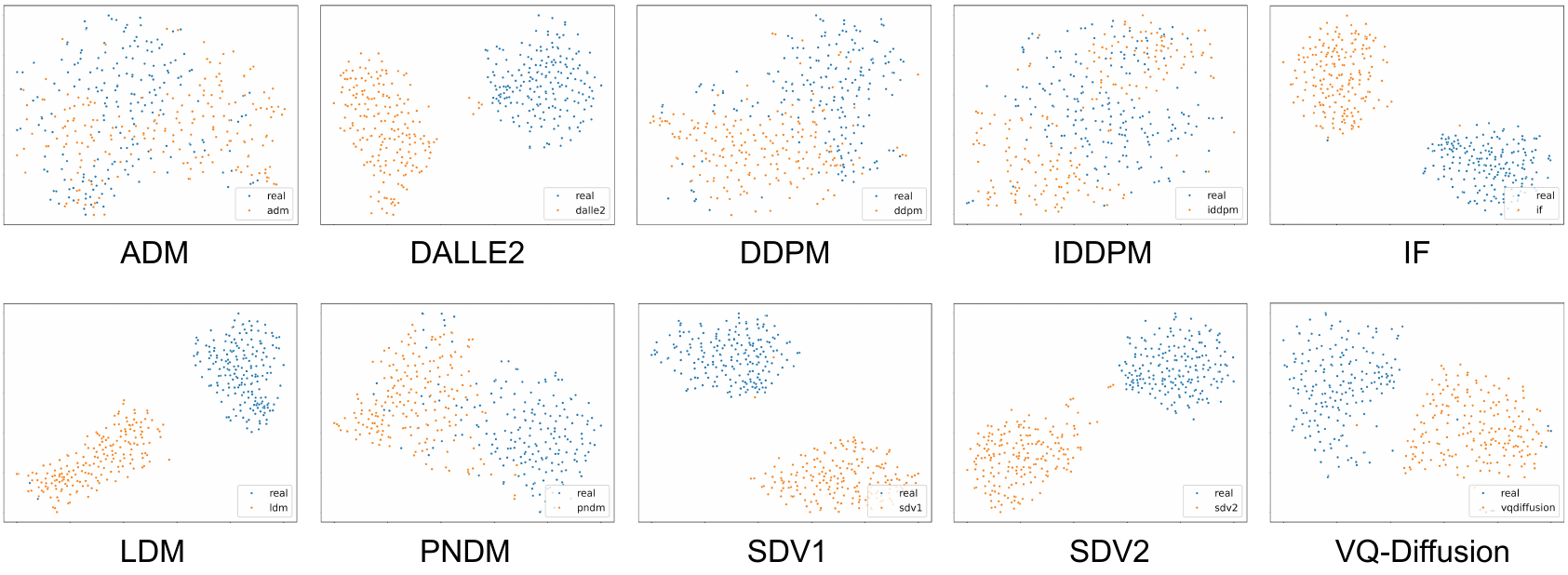}
    \caption{Qualitative result of features extracted by TOFE with Stable Diffusion v1.4.}
    \label{fig:T-SNE_Stable_version_1_4}
\end{figure*}

We use Stable Diffusion v2.0 in the above experiments. Here we make an ablation study of its version to Stable Diffusion v1.4. The quantitative result is in Table~\ref{tab:MMD_JS_Stable_version_1_4} and the qualitative result is in Figure~\ref{fig:T-SNE_Stable_version_1_4}. We can find that in most cases there is a distinction between real and fake distribution.

\begin{table}[htb]
\centering
\caption{Quantitative result of features extracted by TOFE with Stable Diffusion v1.4.}
\resizebox{\linewidth}{!}{
\begin{tabular}{cc|cccccccccc}
\toprule[2pt]
& \textbf{Metric} & \textbf{ADM} & \textbf{DALLE2} & \textbf{DDPM} & \textbf{IDDPM} & \textbf{IF} & \textbf{LDM} & \textbf{PNDM} & \textbf{SD-V1} & \textbf{SD-V2} & \textbf{VQ-Diffusion} \\
\midrule[2pt]
\multirow{2}{*}{} & \textbf{MMD} $\uparrow$ & 0.370 & 2.414 & 0.190 & 0.044 & 6.302 & 0.172 & 2.718 & 6.361 & 1.134 & 4.405 \\
 & \textbf{JS} $\uparrow$ & \num{3.841e-3} & \num{1.932e-2} & \num{6.392e-5} & \num{2.324e-5} & \num{1.353e-1} & \num{3.090e-4} & \num{3.819e-2} & \num{1.463e-1} & \num{3.663e-3} & \num{6.181e-2} \\
\bottomrule[2pt]
\end{tabular}}
\label{tab:MMD_JS_Stable_version_1_4}
\end{table}

\newpage
\subsection{TOFE with $T$ = 50}\label{sec:TOFE_timestep_50}
\begin{figure}[htb]
    \centering
    \includegraphics[width=0.9\linewidth]{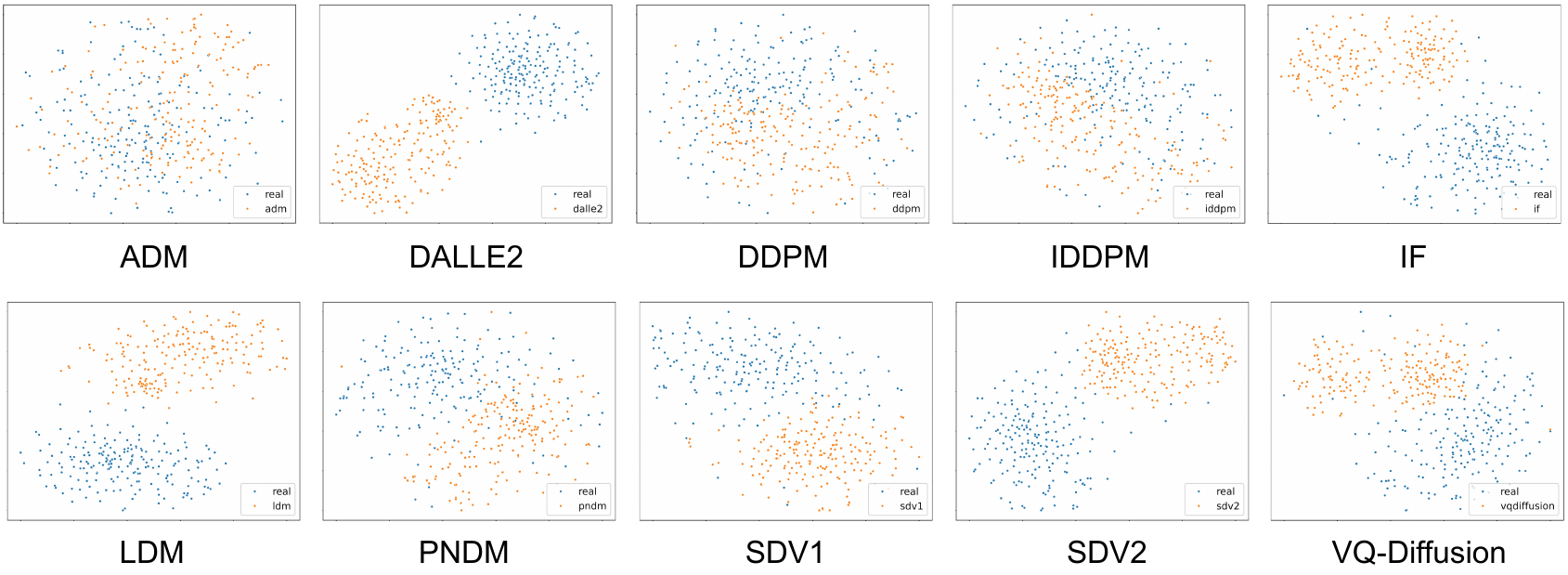}
    \caption{Qualitative result of features extracted by TOFE with $T$ = 50.}
    \label{fig:T-SNE_T_50_t_1}
\end{figure}

Here we demonstrate the representation achieved by TOFE when total timestep $T$=50. There are 50 refined representations and we take the representation at timestep $t$=1 for following qualitative and quantitative analysis. The quantitative result is in Table~\ref{tab:MMD_JS_T_50_t_1} and the qualitative result is in Figure~\ref{fig:T-SNE_T_50_t_1}. We can find that in most cases there is a distinction between the real and fake distribution.
\begin{table}[htb]
\centering
\caption{Quantitative result of features extracted by TOFE with $T$ = 50.}
\resizebox{\linewidth}{!}{
\begin{tabular}{cc|cccccccccc}
\toprule[2pt]
& \textbf{Metric} & \textbf{ADM} & \textbf{DALLE2} & \textbf{DDPM} & \textbf{IDDPM} & \textbf{IF} & \textbf{LDM} & \textbf{PNDM} & \textbf{SD-V1} & \textbf{SD-V2} & \textbf{VQ-Diffusion} \\
\midrule[2pt]
\multirow{2}{*}{} & \textbf{MMD} $\uparrow$ & 0.044 & 1.737 & 0.735 & 0.148 & 4.451 & 1.984 & 2.857 & 3.655 & 0.927 & 3.190 \\
 & \textbf{JS} $\uparrow$ & \num{3.675e-5} & \num{1.115e-2} & \num{8.000e-3} & \num{8.835e-4} & \num{7.252e-2} & \num{1.450e-2} & \num{3.278e-2} & \num{5.588e-2} & \num{5.368e-3} & \num{4.071e-2} \\
\bottomrule[2pt]
\end{tabular}}
\label{tab:MMD_JS_T_50_t_1}
\end{table}

\newpage
\subsection{Corruptions Examples in Robustness Analysis}
Here we give the visualization of corruption and their severity we used in robustness evaluation.
\begin{figure}[htb]
    \centering
    \includegraphics[width=0.9\linewidth]{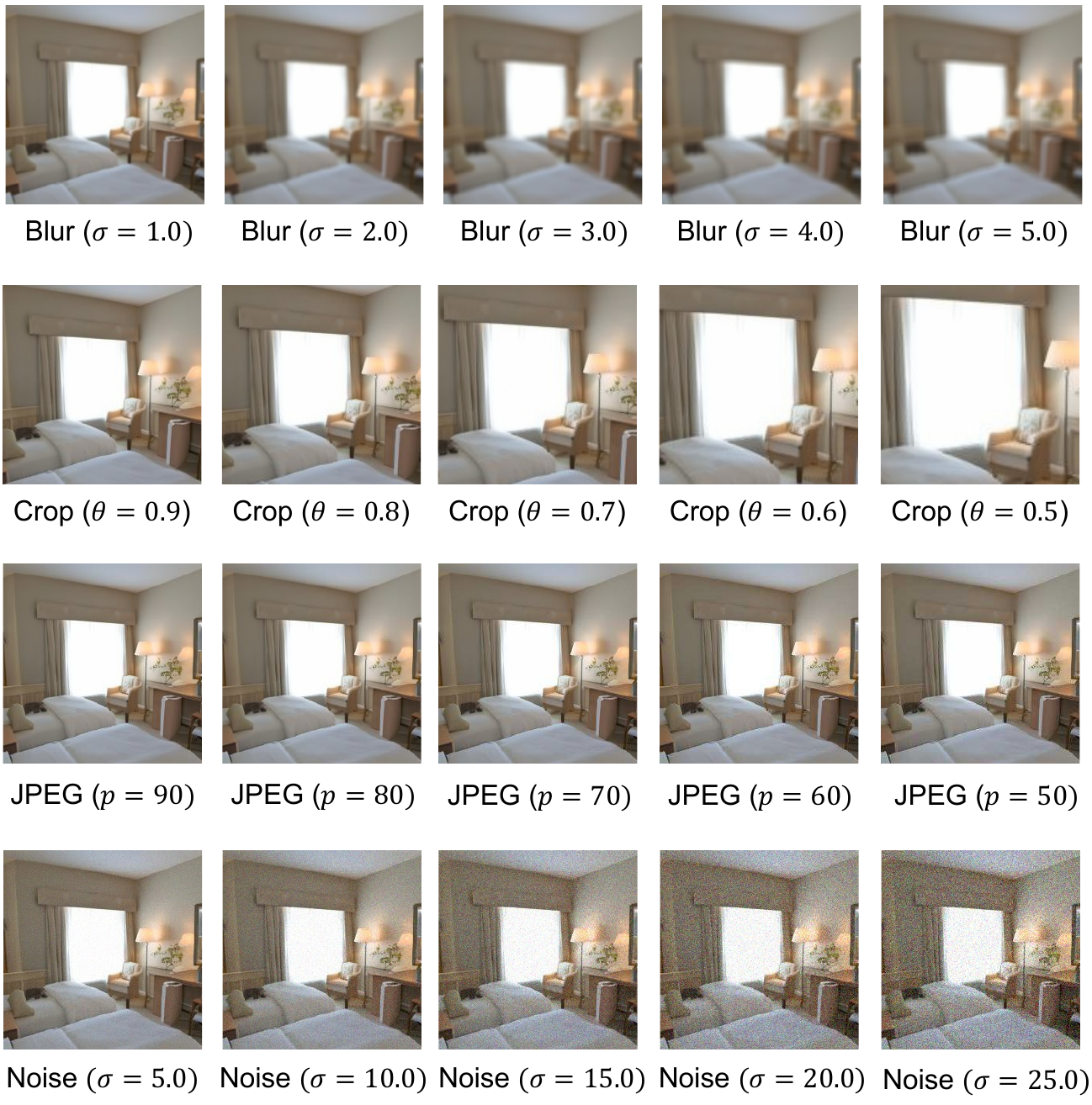}
    \caption{Visualization of different perturbations, including Gaussian noise, Gaussian blur, cropping, and JPEG compression.}
    \label{fig:appendix_robustness_image}
\end{figure}

\end{document}